\documentclass[journal]{IEEEtran}
\usepackage{graphicx} 

\usepackage{tabularray}

\ifCLASSINFOpdf
\else
\fi
\makeatletter
\def\hlinewd#1{%
\noalign{\ifnum0=`}\fi\hrule \@height #1 \futurelet
\reserved@a\@xhline}
\makeatother

\begin{document}
%
\title{Concurrent Classifier Error Detection (CCED) in Large Scale Machine Learning Systems}

%
%
%

\author{
Pedro Reviriego, Ziheng Wang, \'Alvaro Alonso, Zhen Gao, Farzad Niknia, Shanshan Liu and Fabrizio Lombardi
\thanks{
P. Reviriego is with Universidad Polit\'ecnica de Madrid, 28040 Madrid, Spain.  Email: pedro.reviriego@upm.es. \\
Z. Wang is with Northeastern University, Dept. of ECE, Boston, MA 02115, USA. Email: wang.zihe@northeastern.edu. \\
A. Alonso is with Universidad Polit\'ecnica de Madrid, 28040 Madrid, Spain.  Email: alvaro.alonso@upm.es. \\
Z. Gao is with Tianjin University, Tianjin 300072, China. Email: zgao@tju.edu.cn. \\
F. Niknia is with Northeastern University, Dept. of ECE, Boston, MA 02115, USA. Email: niknia.f@northeastern.edu. \\
S. Liu is with New Mexico State University, Klipsch School of ECE, Las Cruces, NM 88003, USA. Email: ssliu@nmsu.edu. \\
F. Lombardi is with Northeastern University, Dept. of ECE, Boston, MA 02115, USA. Email: lombardi@ece.neu.edu. 


}
}

\maketitle

\begin{abstract}
The complexity of Machine Learning (ML) systems increases each year, with current implementations of large language models or text-to-image generators having billions of parameters and requiring billions of arithmetic operations. As these systems are widely utilized, ensuring their reliable operation is becoming a design requirement. Traditional error detection mechanisms introduce circuit or time redundancy that significantly impacts system performance. An alternative is the use of Concurrent Error Detection (CED) schemes that operate in parallel with the system and exploit their properties to detect errors. CED is attractive for large ML systems because it can potentially reduce the cost of error detection.
In this paper, we introduce Concurrent Classifier Error Detection (CCED), a scheme to implement CED in ML systems using a concurrent ML classifier to detect errors. CCED identifies a set of check signals in the main ML system and feed them to the concurrent ML classifier that is trained to detect errors. 
The proposed CCED scheme has been implemented and evaluated on two widely used large-scale ML models: Contrastive Language–Image Pre-training (CLIP) used for image classification and Bidirectional Encoder Representations from Transformers (BERT) used for natural language applications. The results show that more than 95\% of the errors are detected when using a simple Random Forest classifier that is order of magnitude simpler than CLIP or BERT. These results illustrate the potential of CCED to implement error detection in large-scale ML models.

\end{abstract}

\begin{IEEEkeywords}
Machine learning, soft errors, concurrent error detection, CLIP, BERT. 
\end{IEEEkeywords}

%
\IEEEpeerreviewmaketitle

\section{Introduction}

\IEEEPARstart{T}{he} rapid development of Machine Learning (ML) in applications such as computer vision or natural language processing has led to large-scale models with billions of parameters \cite{LLM1}, \cite{TTIM1}. Text-to-image generators such as DALL-E, Stable Diffusion, and MidJourney, or large language models (such as GPT) are now used by millions of people every day and are being widely utilized in many application domains. In particular, large-scale ML systems are being considered for safety-critical applications \cite{ChatGPTrobustness}, \cite{GenAISafetyCritical}. This trend is expected to continue making large-scale ML a fundamental part of safety-critical systems \cite{AI_safety1},\cite{AI_Space},\cite{AI_safety2}.     


Dependability is essential when ML systems are used in safety-critical applications \cite{DepAI1}, \cite{DepAI2}. This requirement has triggered significant efforts to define safety-related evaluation metrics for ML systems \cite{AISafetyCritical1}, \cite{AISafetyCritical2}, \cite{AISafetyCritical3}. Dependability has also been pursued from the fault-tolerance perspective when implementing models, to handle circuit-level errors and faults that can corrupt data integrity \cite{FTAIHW}. Over the years, different error-tolerant techniques have been widely studied for popular ML models, ranging from low-complexity algorithms such as \textit{k} Nearest Neighbors, Random Forests and Support Vector Machines \cite{FTKNNRF}, \cite{FTSVM}, to higher-complexity algorithms such as Neural Networks \cite{FTNN1}, \cite{FTNN2}, \cite{FTNN3}, \cite{FTNN4}. Recently, due to their capability of performing advanced tasks, the error-tolerant design for Deep Neural Networks (DNNs) has also attracted substantial interest, allowing their use in safety-critical applications \cite{FTDNN1}, \cite{FTDNN2}, \cite{FTDNN3}, \cite{FTDNN4}. 

Error detection is often targeted when designing error-tolerant schemes for advanced ML systems. This occurs because error correction inherently is more costly and systems usually have a default safety-mode to handle abnormal situations \cite{SafetyAutomotive}.
Traditional error detection mechanisms mainly rely on circuit or time redundancy, which significantly impacts system performance \cite{SofErrorMitigation}. For example, to detect transient errors, inference can be run twice on the same input data and if the results are different then an error is detected. This, however, has a large performance impact because the speed of the system is reduced to half and the energy consumed twice as much. 

An alternative solution is the use of Concurrent Error Detection (CED) schemes. CED operates in parallel with the ML system and exploits its algorithmic properties to detect errors \cite{CED}. Specifically, CED is more attractive for large-scale ML systems because it can potentially reduce the overhead of error detection and thus, it reduces this burden on the entire system. However, the design of CED for large-scale ML systems faces challenges as many different algorithms are combined, and implementing CED for all of them can be overly complex. A different approach would be to employ an auxiliary ML classifier that operates concurrently with the main ML system checking its outputs to detect errors, thus creating a concurrent classifier CED scheme. To the best of the authors' knowledge, such an approach has not been proposed in the literature. 

In this paper, an efficient error-detection scheme for large-scale ML systems, namely Concurrent Classifier Error Detection (CCED) is presented and evaluated. The principle of CCED is to identify a set of internal check signals in the main ML system and feed them to a concurrent classifier for error detection. This is achieved because that errors that corrupt the output of the ML system tend to create unique patterns on the check signals that are different from those of normal operation. Since the size of the concurrent classifier is very small compared to the main model, CCED can potentially achieve error detection with a negligible overhead. 

The rest of this paper is organized as follows. Section \ref{Preliminaries} covers the preliminaries on large-scale ML models using Contrastive Language-Image Pre-training (CLIP) and Bidirectional Encoder Representations from Transformers (BERT) as examples. Then, the challenges when implementing error detection and in particular, CED on these models are discussed. Section \ref{CCED} presents the proposed CCED scheme describing how it can be applied to CLIP and BERT models. CCED is evaluated in section \ref{Evaluation} in terms of error detection effectiveness and the overhead introduced over the unprotected ML model. Finally, the paper ends with the conclusion in Section \ref{Conclusion}. 

\section{Preliminaries}
\label{Preliminaries}

This section first briefly reviews two widely used large-scale machine learning models to illustrate their distinctive features when compared to classical models. Then, the challenges in implementing CED in these large models are discussed showing the limitations of traditional approaches that try to exploit the features of the algorithms to perform CED. Finally, the error model considered is briefly described.

\subsection{CLIP}
\label{CLIP}

\begin{figure}[h]
	\centering
	\includegraphics[scale=1.05]{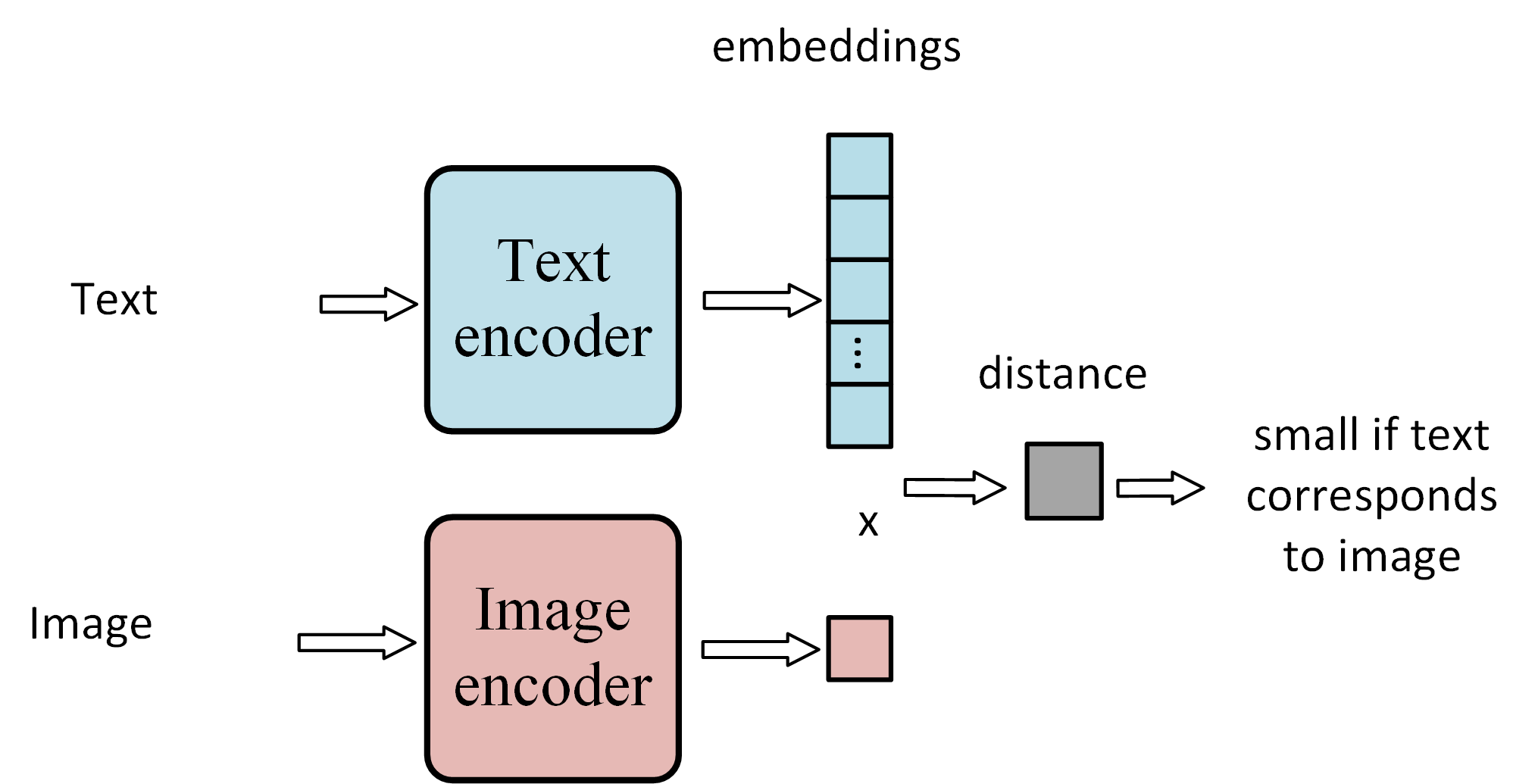}
	\caption{Structure of the CLIP Model}
	\label{fig:CLIP}
\end{figure}

The first large-scale model that we consider is Contrastive Language-Image Pre-training (CLIP) \cite{CLIP} which combines an image encoder and a text encoder as shown in Figure \ref{fig:CLIP}. Both encoders are trained to map to the same embeddings so that images can be associated with the text. Training tries to minimize the distance of the embeddings for images that correspond to the text and to maximize it for images that do not correspond to the text. This enables for example zero-shot learning in which for a given image, the embeddings are computed, and then the distances to several texts, each corresponding to a type of object, are calculated. Finally, the text with the closest embeddings is selected as the classification result. CLIP is also widely used in text-to-image generation in which a pre-trained text encoder is fed with the text prompt to produce embeddings that are then used as input to guide the image generation model \cite{CLIPT2I}.

CLIP typically uses large neural networks for both text and image encoders. For example, deep neural networks like the 50-layer ResNet \cite{RN50} or vision transformers \cite{VIT} such as ViT-L/14. This leads to a very large number of parameters and a complex system. 

\begin{figure}[h]
	\centering
	\includegraphics[scale=0.9]{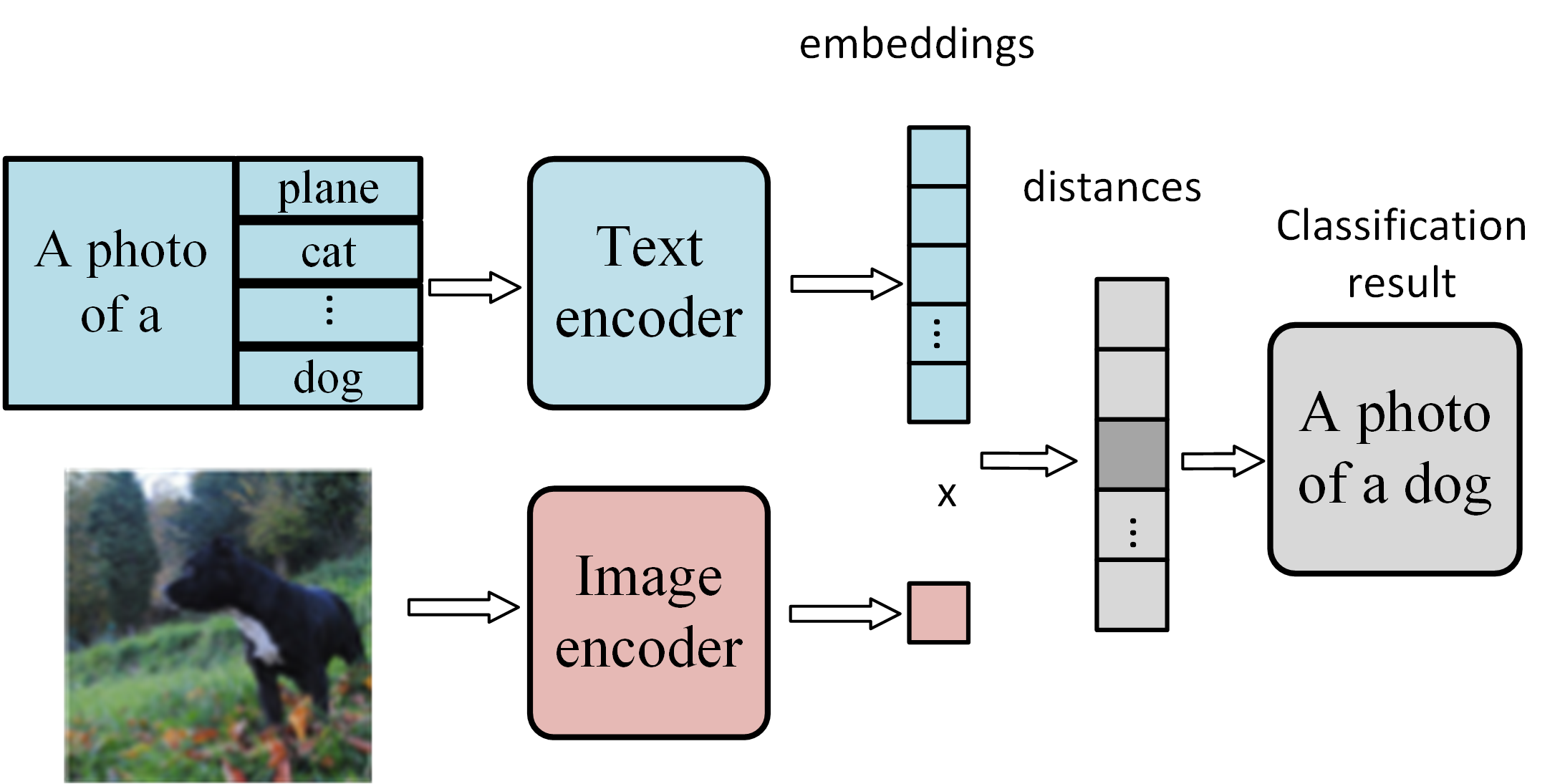}
	\caption{CLIP used for zero-shot learning}
	\label{fig:CLIPshot}
\end{figure}

In our evaluation, we use CLIP for zero-shot image classification where the embeddings of different texts corresponding to the classes are compared to the embeddings of the image to be classified. The one with the shortest cosine distance is selected as the class for the input image. This is illustrated in Figure \ref{fig:CLIPshot}.

\subsection{BERT}
\label{BERT}

\begin{figure}[t]
	\centering
	\includegraphics[scale=0.4]{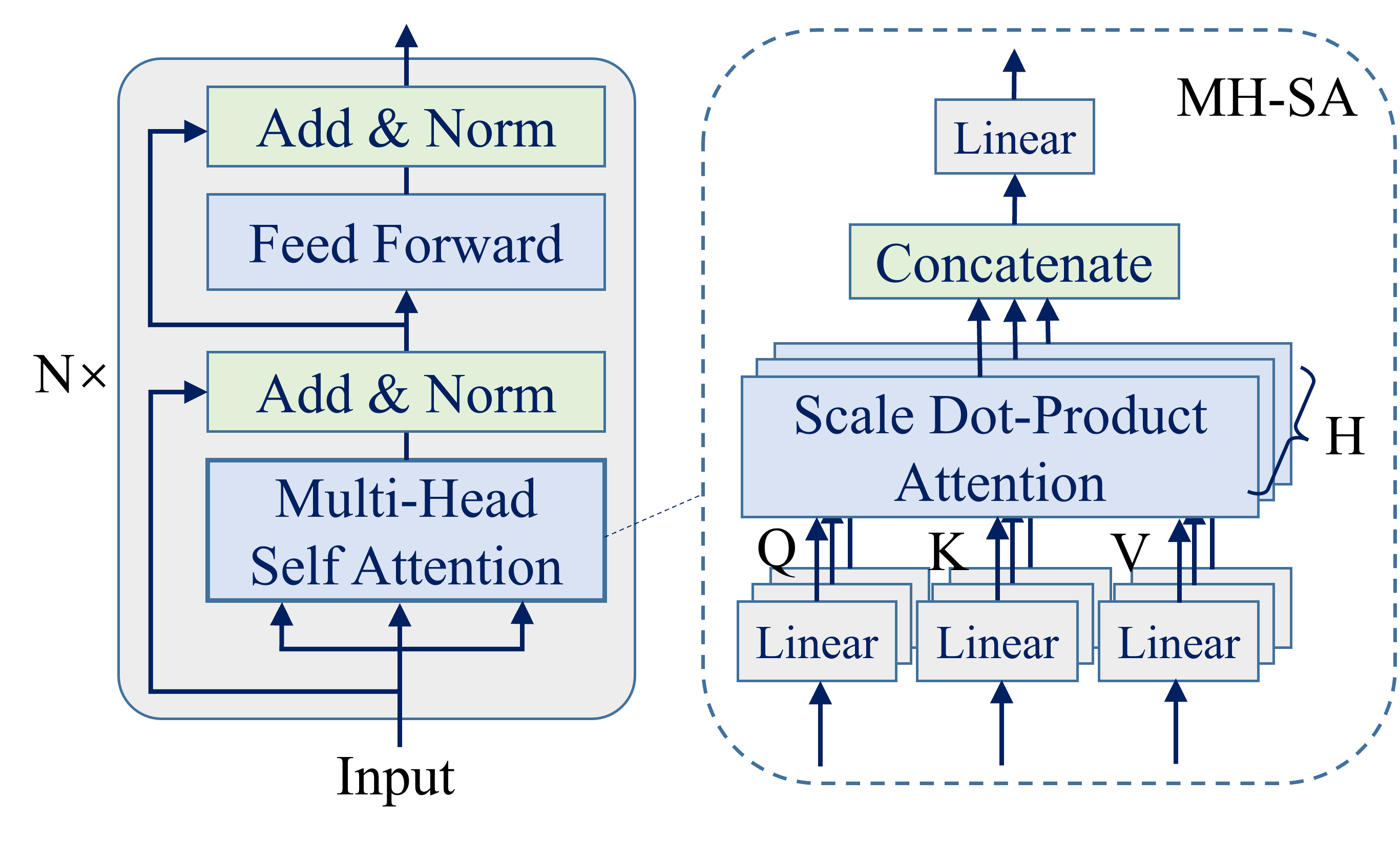}
	\caption{Structure of the BERT Model: structure (left), detail of the SA block (right).}
	\label{fig:BERT}
\end{figure}

The second model used to illustrate large-scale ML systems is Bidirectional Encoder Representations from Transformers (BERT) \cite{BERT} that is used in Natural Language Processing (NLP) applications. As illustrated in Figure \ref{fig:BERT}, the BERT model is composed of a stack of N Transformer encoders, and each encoder consists of a multi-head Self-Attention (SA) layer and a Feed-Forward Network (FFN). Each head of SA module performs scaled dot-product attention for a query matrix, a key matrix, and a value matrix. The FFN is composed of two fully-connected layers and a non-linear activation operation between them. The standard BERT model includes 12 encoders (N = 12), and each MH-SA includes 12 heads (H = 12). So it is a very large model with 110 million parameters.

\begin{figure}[t]
	\centering
	\includegraphics[scale=0.5]{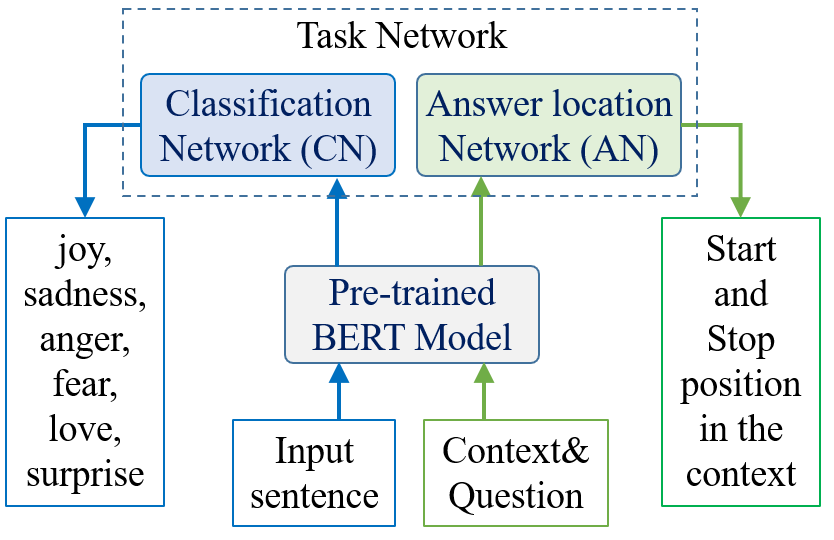}
	\caption{Stack of BERT model and task network for different NLP applications}
	\label{fig:BERT+Task}
\end{figure}

BERT is pre-trained using specific tasks such as mask language modeling and next-sentence prediction to extract embeddings of the text input which can then be used for different NLP tasks such as for example, to infer the emotion from a text \cite{Emotion} or to find the answer to a question in a text. This is typically done by connecting a neural network to the outputs of BERT to perform the task at hand. In the case of emotion analysis, this can be a shallow neural network that is trained to classify emotions. This is illustrated in Figure \ref{fig:BERT+Task} for both use cases. 

\subsection{Challenges for error detection in large-scale ML models}
\label{Challenges}

Implementing error detection in large-scale ML systems poses new challenges when compared to simpler models. The first one is that the huge number of parameters and operations that are needed in those systems leave very little room to add redundancy. Therefore, error detection must be implemented with a very small fraction of the cost of the original system. This characteristic rules out the use of most traditional time and space redundancy-based schemes such as duplication with comparison or recomputation and comparison. Additionally, as the complexity of the ML systems increases in each new release or generation, the error detection approach should be scalable so that it can be applied to more complex models with a similar cost in relative terms. Finally, and for the same reason, the error detection scheme should be generic so that it can be integrated as part of the ML system design flow and does not require a complete redesign when the ML model changes. Therefore, ideally, error detection schemes should: 

\begin{enumerate}
    \item Have a small relative cost.
    \item Be scalable.
    \item Be independent of the details of the ML system model used.
\end{enumerate}

The  first requirement points to the use of CED so reusing the ML system properties to detect errors. However, given the complexity of large-scale ML systems (that use a variety of components and algorithms), the implementation of CED is not straightforward. One possibility could be to try to implement CED using existing techniques for each of the model components at some granularity. For example, implement CED for the Neural Network blocks, for the matrix multiplications, and so forth. This approach however leads to a complex design process in which different CED techniques have to be used for each block. Even worse, when the main ML model changes, which happens frequently as models are improving continuously, the CED must be redesigned. In summary, using CED at the block level does not meet the requirements of scalability and independence from the main ML model. Therefore, there is a need for new CED schemes that can meet the requirements of large-scale ML systems.

\subsection{Error model}
\label{ErrorModel}


The error model considered in this paper consists of transient errors that correspond for example to radiation-induced soft erros as major issue in modern computing systems \cite{SoftErrors}. Therefore, we consider errors in the machine learning model parameters when they are used to perform inference. In many cases, the parameters will be loaded to perform inference from the main memory or storage for each inference as they cannot all be stored on-chip. Therefore, the errors are assumed to be soft errors that affect the computation of an inference but they do not have a persistent effect; if we run the inference again, the parameters will be loaded again and the error will be no longer present. Therefore, for this scenario, if CED is implemented when an error is detected, it can be corrected by running the inference a second time.

\section{Concurrent Classifier Error Detection (CCED)}
\label{CCED}

In this section, we present Concurrent Classifier Error Detection (CCED), a scheme to detect errors in large-scale ML systems that addresses the challenges discussed in section \ref{Challenges}, so providing low-cost, scalable, and generic CED for large-scale ML systems. Initially, the overall principle and approach are presented; then we consider the selection process of the inputs to the concurrent classifier and finally discuss the implementation of the proposed scheme.

\subsection{Overall approach}
\label{Approach}

The proposed CED scheme is based on the following observations:

\begin{enumerate}
    \item We are only interested in detecting errors that change the classification result. The rest of the errors do not have an impact on the outcome of the ML system and since they are not persistent, they can be ignored.
    \item The errors that do change the classification result would be expected to introduce in most cases significant changes in the values of some of the nodes of the ML system; hence they take values that are different from those during normal operation.
    \item A classifier can, from the previous observation, be trained with the values of the nodes in the error-free and error cases to detect errors. This classifier can then operate concurrently with the main ML system to perform error detection.
    \item The cost of such a classifier should be negligible compared to a large-scale ML system because it is much simpler.
\end{enumerate}

\begin{figure*}[t]
	\centering
	\includegraphics[scale=0.8]{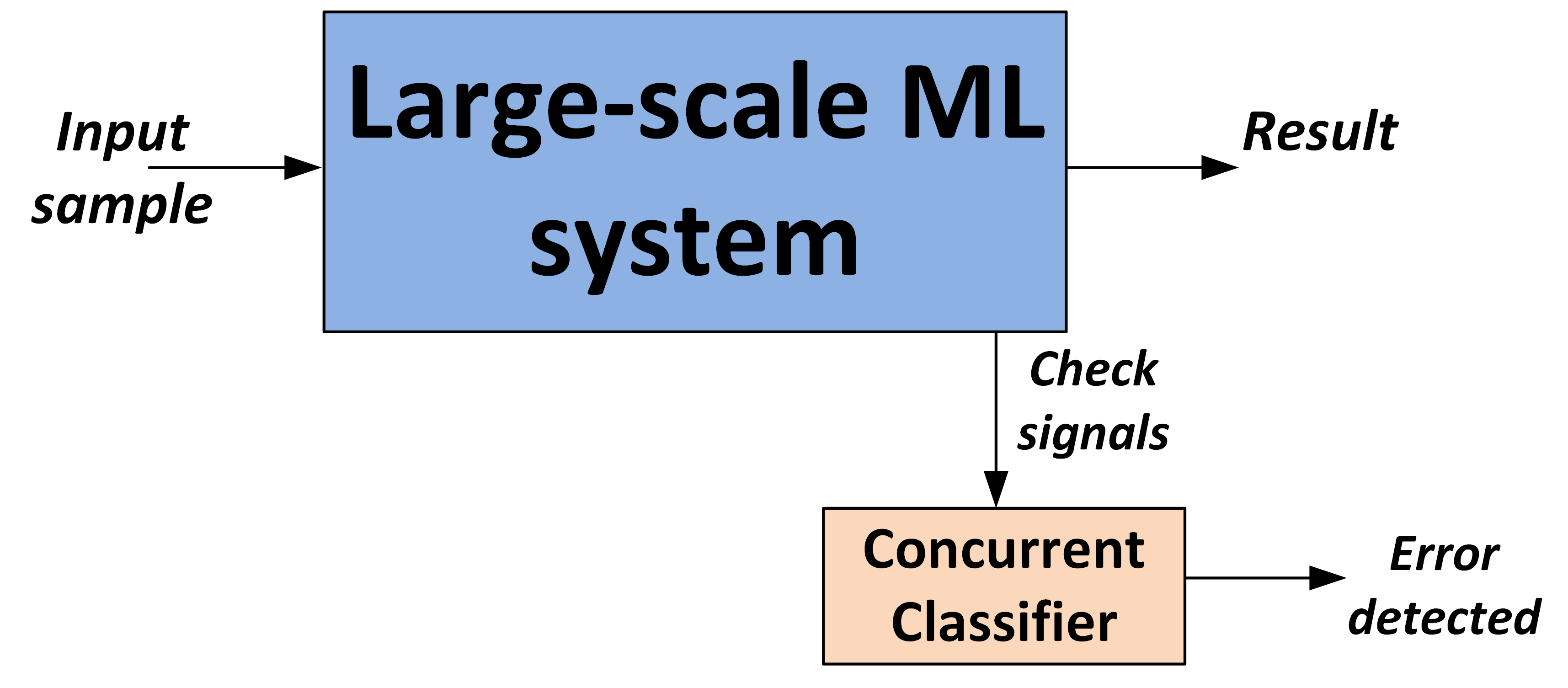}
	\caption{Proposed Concurrent Classifier Error Detector (CCED) scheme.}
	\label{fig:CCED}
\end{figure*}

From those observations, a general scheme to implement CED in large-scale ML systems can be to use a small classifier that checks some internal nodes of the ML system to detect errors. Therefore, this is a Concurrent Classifier Error Detector (CCED). The overall principle is illustrated in Figure \ref{fig:CCED}, the concurrent classifier takes as inputs the values of some of the nodes of the main ML system and uses them to detect errors. This approach addresses the three requirements discussed in section \ref{Challenges}. First, it introduces a very small overhead as the concurrent classifier can be simple, for example, a Random Forest, as discussed in section \ref{CC}, i.e., its cost would therefore be negligible when compared to the main ML model. For the same reasons it would be scalable as the complexity of the concurrent classifier does not depend on the size or complexity of the main model but only on the data patterns introduced by the errors on the nodes being monitored. The approach is also independent of the implementation details of the main ML system as it only needs to take the values from it as a black-box. Finally, from a design perspective, it integrates nicely with the main ML system design as now the CED is another (much simpler) ML problem which fits naturally in the design flow. Therefore, the proposed CCED has the potential to address the challenges of implementing error detection in large-scale ML systems. However, the first step is to check that the outputs of the system with and without errors have different patterns that can be separated by a simple classifier. This will be confirmed by the results presented in the evaluation section for two well-known ML models: CLIP and BERT. 

\begin{figure}[t]
	\centering
	\includegraphics[scale=0.6]{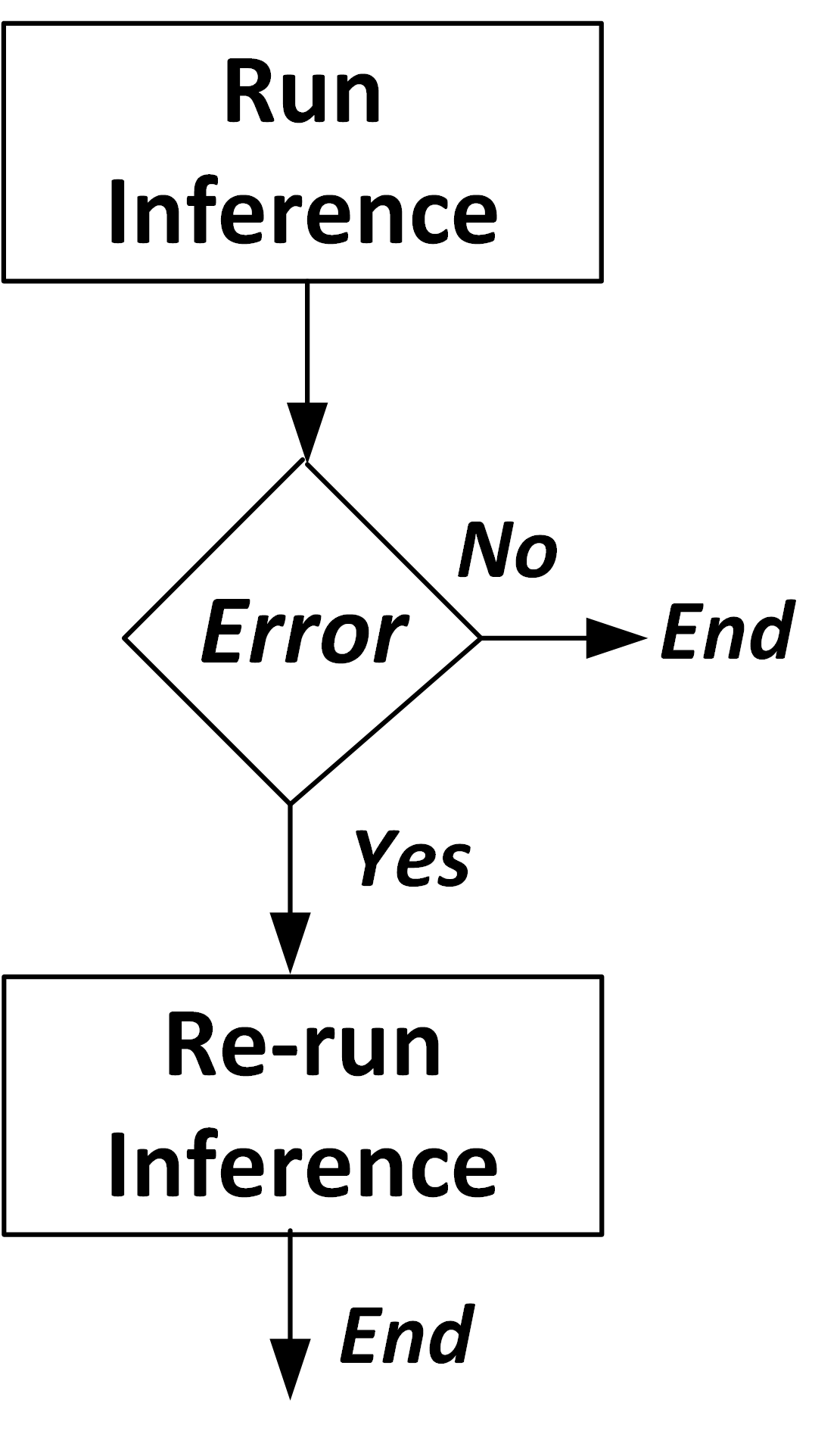}
	\caption{Operation of the proposed CCED scheme.}
	\label{fig:CCEDoperation}
\end{figure}

CCED upon detecting an error will re-run the inference such that if the error was due to a soft error, it will no longer be present, and the second run would be correct. Instead, if the error was due to a miss classification, the concurrent classifier will also detect an error in the second run that is ignored. The overall process is illustrated in Figure \ref{fig:CCEDoperation}. 

The performance of CCED depends on two variables: the number of false negatives and false positives. A false negative occurs when the concurrent classifier fails to identify an inference with an error. Instead, a false positive occurs when the concurrent classifier detects an error when there is none. False positives would trigger additional inferences to correct errors that do not exist while false negatives will leave the errors undetected. Therefore, their impact on performance is qualitatively different and the decision threshold used in the concurrent classifier can be used to make trade-offs between the two. For example, to maximize the error detection rate (minimize false negatives) subject to a given percentage of re-computations (false positives). 
This models a system that can afford a given fraction of redundant re-computations to detect errors, for example, 10\%, and given that constraint wants to maximize error detection. This will be further discussed in the evaluation section.

\subsection{Selection of the nodes to monitor}
\label{Nodes}

In principle, any node in the main ML system can be monitored by the concurrent classifier. A simple approach would be to select the nodes just before the outputs. For example, in the case studies considered in this work, namely CLIP for zero-shot learning and BERT for emotion detection and question and answer, the softmax values used to determine the final class selected can be used for monitoring. This has several advantages, first, it makes the selection of the nodes straightforward for classification systems; second, by nature, the number of nodes is small which reduces the complexity of the concurrent classifier and third any error that has an effect on the system must affect these nodes and thus can potentially be detected using the values of these nodes. Therefore, next, these nodes are used for monitoring and are the inputs to the concurrent classifier.   

The use of additional and/or alternative nodes is left for future work and can be used to improve the error detection rate or to reduce the false positive rate. The nodes can be selected for example by checking the impact of errors on the different nodes and choosing for monitoring the ones that have the largest differences. As this additional process can only improve the error detection rate, the results presented in the evaluation section are a lower bound on the detection rates that can be achieved by the proposed CCED technique. 

\subsection{Concurrent classifier implementation}
\label{CC}

One of the assumptions behind our proposed CCED scheme is that errors can be detected by monitoring a small set of nodes and using a simple classifier. Therefore, the concurrent classifier can be implemented using classical machine learning algorithms such as Logistic Regression (LR), Support Vector Machines (SVM), or Random Forest (RF) which in simple classification problems can achieve an accuracy like that of more complex models such as deep neural networks \cite{SimpleML}. Therefore, in the proposed design, we consider such simple classifiers and leave the use of more complex ML models for the concurrent classifier for future work. As for the selection of the monitoring nodes, this puts us in a worst-case in terms of the error detection rate that can be achieved. Better error detection may be obtained by using more complex CCs at the cost of additional overheads. 

\section{Evaluation}
\label{Evaluation}

To assess the proposed CCED scheme over a wide range of scenarios, it has been evaluated using CLIP for zero-shot classification on several image datasets and with BERT for sentiment analysis and questions and answers. The following subsections describe the case studies and the methodology employed in the experiments and then results are presented. 

\subsection{Case studies}

\subsubsection{CLIP}

For CLIP, three widely used datasets: CIFAR10 \cite{CIFAR10}, CIFAR100 \cite{CIFAR100} and mini-imagenet \cite{MiniImage} are used as inputs.  The first one has 10 classes and the second and third 100. The CLIP network is configured using two encoders with different complexity and performance: the simpler pre-trained model “RN50” (modified 50-layer ResNet) and the more advanced "ViT-L/14" (a vision transformer). Classification is performed using softmax and the label with the largest value is selected as the final prediction. The accuracy achieved by CLIP is summarized in Table \ref{tab:accuracy_clip}, it can be seen that the "ViT-L/14" model achieves significantly better performance. This comes at a cost because the two models have respectively  102 million parameters for RN50 and 427 million for "ViT-L/14".  Among the datasets, the accuracy is higher for CIFAR10 because it is an easier data set with fewer classes.

\subsubsection{BERT}

For BERT, sentence emotion classification and question answering are used as case studies. In the first one, CLIP must select one of six emotions for each sentence. The BERT model is extended with a classification network (CN) with two fully-connection layers that generate six scores, one per emotion. The stacked model (pre-trained standard BERT plus CN) from Huggingface\footnote{https://huggingface.co/Vasanth/bert-base-uncased-finetuned-emotion.} is used in our evaluation. In the second case, BERT is extended with an answer location network (AN) with two independent linear transformations, one to detect the start of the answer and the other to detect the end of the answer. Again the stacked model (pre-trained standard BERT plus AN) from Huggingface\footnote{https://huggingface.co/phiyodr/bert-base-finetuned-squad2.} is used in our evaluation. The accuracy achieved by BERT in both tasks is given in table \ref{tab:accuracy_bert}.

\begin{table}[]
\centering
\caption{Accuracy of CLIP in the case studies considered (Top-1 accuracy)}
\label{tab:accuracy_clip}
\begin{tabular}{|c|c|c|c|}
\hline
 Model   &  CIFAR10 & CIFAR100 &  mini-imagenet   \\  \hlinewd{1pt}
 CLIP-RN50     & 68.69\%  & 38.99\% & 68.13\%    \\  \hline
 CLIP-ViT    &95.29\%   &  73.28\% & 84.36\%   \\  \hline
 
 \end{tabular}
\end{table}

\begin{table}[]
\centering
\caption{Accuracy of BERT in the case studies considered}
\label{tab:accuracy_bert}
\begin{tabular}{|c|c|}
\hline
 Case study & Accuracy  \\  \hlinewd{1pt}
 Emotion Classification   &  92.10\%   \\  \hline
 Question Answering   &  70.39\%   \\  \hline
 \end{tabular}
\end{table}

\subsection{Methodolody}

\subsubsection{Dataset creation}

The evaluation methodology starts by generating a data set to train the concurrent classifier. This is done by running inference with and without errors to produce samples of the signals that are used as inputs to the concurrent classifier. For CLIP, first, inference is run with no errors, and the values of the softmax output for the 10 (100) classes are stored for CIFAR10 (CIFAR100 and mini-imagenet). Then a random bit is flipped in one of the parameters of the CLIP model and inference is run again, if the classification result changes, then the softmax values are stored. Using this procedure, a balanced dataset with 10,000 samples with errors that changed the classification result and 10,000 error free samples is built for CIFAR10, CIFAR100, and mini-imagenet. These are the datasets used to evaluate the performance of the concurrent classifier.

For BERT and emotion analysis, a similar procedure is used by storing in this case the six softmax values for both error-free inferences on each element of the test set and then for runs with errors that lead to a change in the output of the classifier. In this case, the dataset has 4000 samples, half with an error and half without errors. This dataset is used to evaluate the concurrent classifier. For question answering, a similar procedure is used but this time, the probability for each word in the context being the start and end of the answer are saved. This means that the number of values is significantly larger than six, it is around 100 $\sim$ 200. The size of the data set for question answering is also 4000 samples. 

\subsubsection{Concurrent classifier training}

The second step in the evaluation is to use the datasets generated in the first step to train a concurrent classifier. A simple Random Forest classifier has been used in the experiments. The rationale as in previous choices for the nodes monitored is to obtain a lower bound on the performance of CCED that can be improved by using more complex classifiers or automatically generated ensembles \cite{AutoML}. For the same reason, the default hyper-parameter values of the library used (sklearn) are not modified to improve performance, our objective is to show that even a simple classifier with default parameters can detect most errors. 


\subsubsection{Performance evaluation}

To evaluate the performance of CCED we fix a percentage of false positives and adjust the decision threshold of the classifier to minimize false negatives given that constraint. As discussed prior to this model, a system can afford a given fraction of re-computations to detect errors. We consider percentages of 5,10,15, and 20\%  as reasonable overhead values in terms of re-computation effort (much lower than recomputing every inference to detect errors when there is no CED). 

The concurrent classifier is trained normally using a subset of the dataset. Then the rest of the dataset is used as the test set to measure the false positives and negatives and also get the classification scores. Finally, the decision threshold is shifted until the maximum number of allowed re-computations is used and the percentage of detected errors on the test set is computed for that threshold.  

\subsection{Detection and re-computation}

For CLIP, the results obtained for the three concurrent classifiers are summarized in Tables \ref{tab:CIFAR10},\ref{tab:CIFAR100},\ref{tab:miniimagenet}. The tables show the percentage of errors detected, the re-computations needed when using the default classifier threshold and then the percentage of errors detected when the threshold is adjusted to have a given percentage of re-computations. 

\begin{table*}[]
\centering
\caption{Error detection rates for different percentages of re-computation for CLIP on CIFAR10 }
\label{tab:CIFAR10}
\begin{tabular}{|c|c|c|c|c|c|}
\hline
 Main model & default (re-comp)  & 5\% & 10\% & 15\%  & 20\%   \\ \hlinewd{1pt}
 RN50 & 94.9\% (8.1\%) & 93.3\% & 95.4\% & 96.9\% & 97.5\% \\  \hline
 ViT & 98.4\% (2.3\%) & 99.4\% & 99.6\% & 99.7\% & 99.8\% \\  \hline
 \end{tabular}
\end{table*}

\begin{table*}[]
\centering
\caption{Error detection rates for different percentages of re-computation for CLIP on CIFAR100}
\label{tab:CIFAR100}
\begin{tabular}{|c|c|c|c|c|c|}
\hline
 Main model & default (re-comp)  & 5\% & 10\% & 15\%  & 20\%   \\  \hlinewd{1pt}
 RN50 & 80.9\% (13.7\%) & 70.3\% & 78.4\% & 81.9\% & 85.0\% \\  \hline
 ViT & 97.5\% (2.0\%) & 98.1\% & 98.7\% & 99.0\% & 99.3\% \\  \hlinewd{1pt}
 \end{tabular}
\end{table*}

\begin{table*}[]
\centering
\caption{Error detection rates for different  percentages of re-computation for CLIP on mini-imagenet}
\label{tab:miniimagenet}
\begin{tabular}{|c|c|c|c|c|c|}
\hline
 Main model   & default (re-comp)  & 5\% & 10\% & 15\%  & 20\%   \\  \hlinewd{1pt}
 RN50  & 70.9\% (13.1\%) & 59.7\% & 67.8\% & 72.3\% & 76.3\% \\  \hline
 ViT & 92.4\% (2.6\%) & 94.2\% & 95.3\% & 95.6\% & 96.1\% \\  \hlinewd{1pt}
 \end{tabular}
\end{table*}

It can be observed that the random forest concurrent classifier provides good results in most cases. Therefore, it seems that random forest can be a good choice for the concurrent classifier\footnote{Note that as discussed before we are taking the random forest classifier with its default settings so it may be possible to achieve better results by selecting other values of the hyper-parameters.}. Focusing on the results, the random forest classifier achieves high detection rates. More than 95\% of the errors are  detected for all three datasets when the ViT network is used for encoder with only 10\% of recomputations and for CIFAR10, the value is over 99\%. Instead, when the simpler RN50 network is used, detection is above 90\% only for the CIFAR10 dataset. The difference may be linked to the accuracy achieved by the main classifier which is also better when using ViT than RN50. This suggests that the proposed CCED scheme works better when the main classifier is also performing well.

\begin{figure}[h]
	\centering
	\includegraphics[scale=0.55]{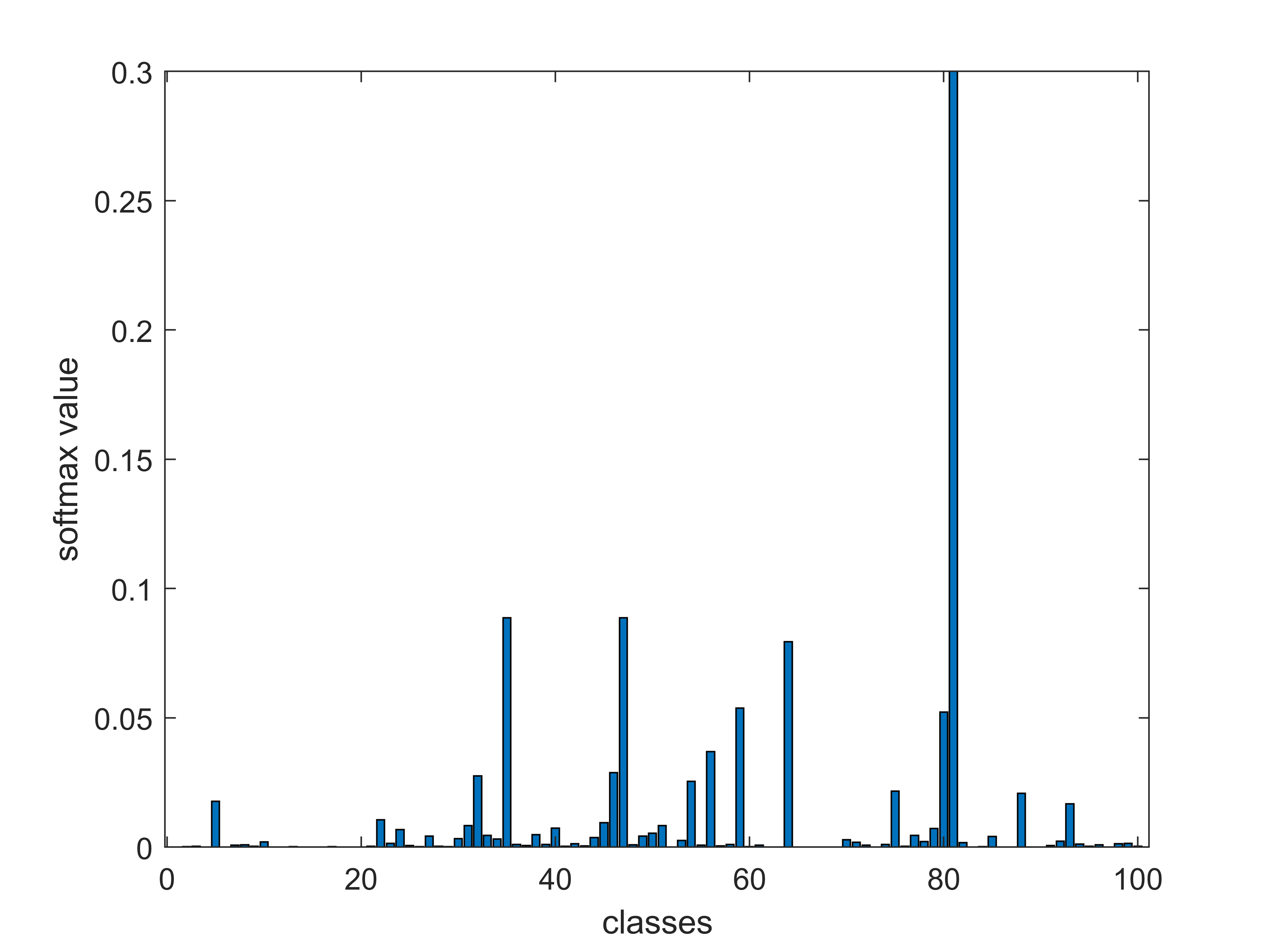}
	\caption{Example of an error-free pattern for CLIP with RN50 and CIFAR100.}
	\label{fig:Pattern100}
\end{figure}
\begin{figure}[t]
	\centering
	\includegraphics[scale=0.55]{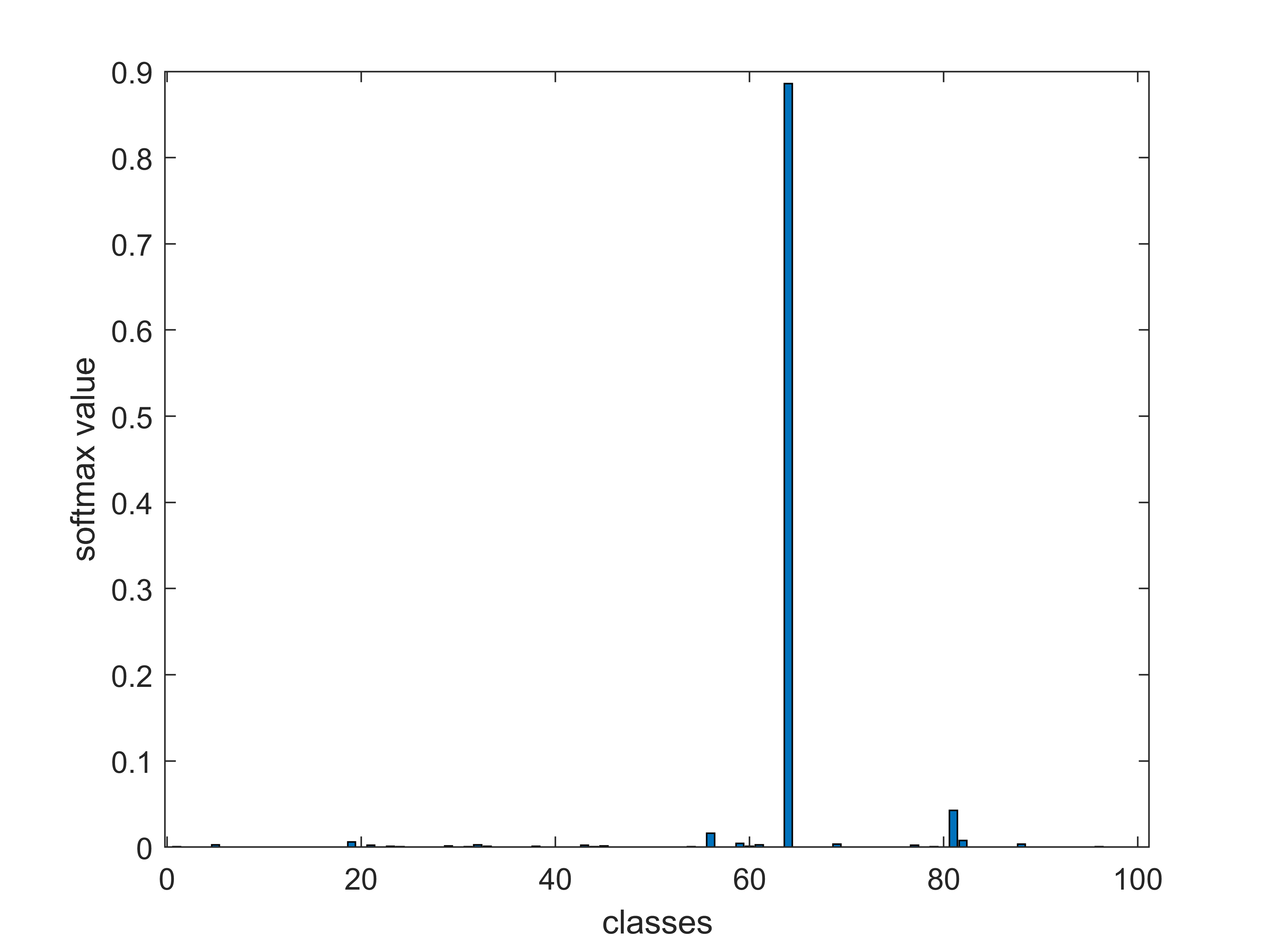}
	\caption{Example of an error-free pattern for CLIP with ViT on CIFAR100.}
	\label{fig:PatternVIT}
\end{figure}

This can be explained as when the main system has good performance, the check signals tend to have clean patterns, for example with one of the classes taking a large value and the remaining much smaller ones. This makes it easier to separate those patterns from the ones induced when an error is inserted in the system which tends to produce noisy patterns. Instead, when the main system has poor performance the error-free patterns are less clean and thus it is harder to differentiate from the ones when an error occurs. Therefore, the performance of the proposed CCED is better when the main ML system has good accuracy. This is illustrated in Figures \ref{fig:Patterns10},\ref{fig:Pattern100} that show typical patterns of the softmax values (which are the inputs to the concurrent classifier) for CIFAR10 and CIFAR100 respectively when using RN50 for the encoder. In the first case, the error-free pattern is better defined and thus it is easier to distinguish from the patterns obtained when there is an error in one of the system parameters. Figure \ref{fig:PatternVIT} shows a typical pattern for CIFAR100 but when ViT is used for the encoder. In this case, the pattern is also cleaner which can be related to the higher accuracy obtained by CLIP in this case and with the better performance of the proposed CCED approach.

\begin{figure}[h]
	\centering
	\includegraphics[scale=0.55]{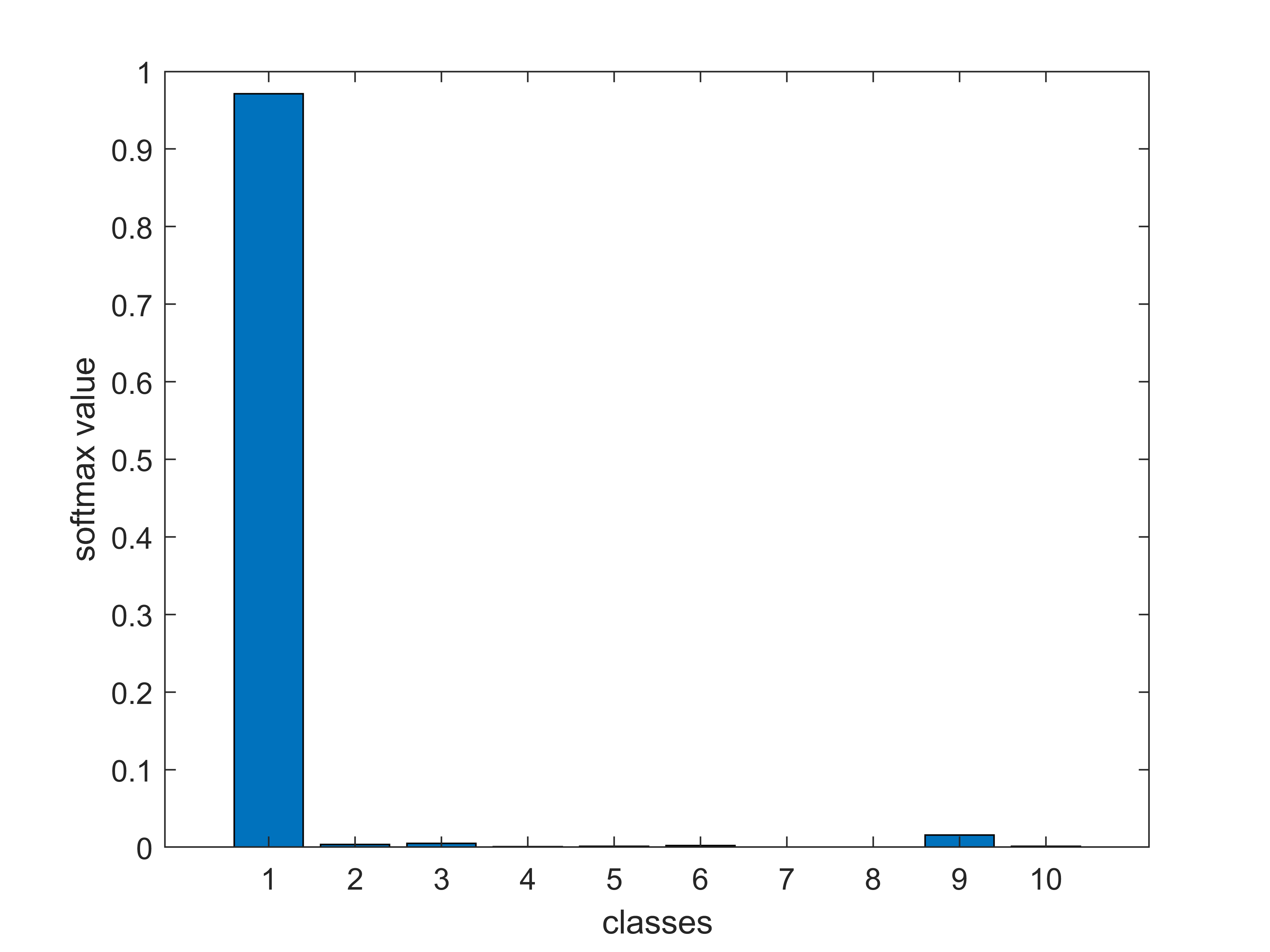}
	\caption{Example of an error-free pattern for CLIP with RN50 and CIFAR10.}
	\label{fig:Pattern10}
\end{figure}

For BERT, when used for emotion classification, most errors can be detected by the concurrent classifier even when the percentage of re-computations is low as shown in the results in Table \ref{tab:BERTemotion}. For example by allowing 10\% of re-computations over 99\% of the errors are detected. The results for question answering are illustrated in Table \ref{tab:BERTquestion} and show similar trends, and again with  10\% of re-computations the concurrent classifier is able to detect over 99\% of the errors. Therefore, the proposed scheme can detect most errors in BERT with a low overhead. In fact, for BERT, the percentage of recomputations can be reduced to 5\%, and still more than 98\% of the errors are detected.

\begin{table}[]
\centering
\caption{Error detection rates for different  percentages of re-computation for BERT in emotion classification}
\label{tab:BERTemotion}
\begin{tabular}{|c|c|c|c|c|}
\hline
 default (re-comp)  & 5\% & 10\% & 15\%  & 20\%   \\  \hlinewd{1pt}
 97.0\% (5.2\%) & 96.9\% & 99.3\% & 99.6\% & 99.9\% \\  \hline
 \end{tabular}
\end{table}

\begin{table}[]
\centering
\caption{Error detection rates for different  percentages of re-computation for BERT in question answering}
\label{tab:BERTquestion}
\begin{tabular}{|c|c|c|c|c|}
\hline
 default (re-comp)  & 5\% & 10\% & 15\%  & 20\%   \\  \hlinewd{1pt}
 98.2\% (2.5\%) & 98.7\% & 99.3\% & 99.6\% & 99.8\% \\  \hline 
 \end{tabular}
\end{table}

\subsection{Complexity}

Finally, it is of interest to compare the complexity of the main ML model with the concurrent classifier. To that end, we use the execution time needed to perform an inference on average as an indication of the complexity. In more detail, the inference is run for 1,000 samples, and the average time is computed. Table \ref{tab:runtime_clip} compares the execution time of the models for CLIP and Table \ref{tab:runtime_bert} for BERT. It can be observed that in all cases, the concurrent classifier is orders of magnitude faster than the main model. The CLIP model takes more than 4000 times to execute than the random forest concurrent classifier. Therefore, the complexity of the concurrent classifier is negligible. For BERT, the ratio is a bit smaller but it is still above 1200 times in both emotion detection and question and answer. The difference between CLIP and BERT is mostly due to the larger runtime of CLIP compared to BERT. 

These results confirm that the concurrent classifier introduces a very small overhead to the system and that the main cost of the proposed scheme is the re-computations induced by false positive error detections of the concurrent classifier.




\begin{table}[]
\centering
\caption{Average run time for CLIP (milliseconds)}
\label{tab:runtime_clip}
\begin{tabular}{|c|c|c|c|}
\hline
 Model   &  CIFAR10 & CIFAR100 &  mini-imagenet   \\  \hlinewd{1pt}
 CLIP-RN50   & 42.7 & 102.4 & 119.4    \\  \hline
 CLIP-ViT    & 101.9  & 233.8 & 251.0   \\  \hlinewd{1pt}
 RF (concurrent)    & 0.0088 &  0.0099 & 0.0100   \\  \hline
 
 \end{tabular}
\end{table}

\begin{table}[]
\centering
\caption{Average run time for BERT (milliseconds)}
\label{tab:runtime_bert}
\begin{tabular}{|c|c|c|}
\hline
 Case study & Emotion Classification   &  Question Answering \\  \hlinewd{1pt}
 BERT     & 8.4 &  10.8   \\  \hlinewd{1pt}
 RF (concurrent)   & 0.0060 & 0.0082    \\  \hline
 \end{tabular}
\end{table}

\section{Conclusion}
\label{Conclusion}
This paper has presented Concurrent Classifier Error Detection (CCED) a scheme to detect errors in large-scale machine learning systems. The proposed method uses a simple classifier that takes as input the values of a small set of nodes of the main system and operates concurrently with it to detect errors. This enables scalable and low-cost error detection that is independent of the main machine learning model implementation and related details. Furthermore, the proposed CCED integrates naturally into the machine learning design flow as error detection is done also using machine learning. The proposed scheme has been evaluated on two widely used large-scale machine learning models. The results show that CCED can detect over 95\% of the errors with only 10\% of recomputations even when using a simple classifier such as a random forest. Therefore, CCED is a promising approach for implementing concurrent error detection in large-scale machine learning systems.  
Future work will explore the selection of the nodes used as inputs to the concurrent classifier and the optimization of the concurrent classifier itself to further increase the error detection rate and reduce the re-computations needed.


%

\ifCLASSOPTIONcaptionsoff
  \newpage
\fi



%

\bibliographystyle{IEEEtran}
\bibliography{cced}

%







\end{document}